\pdfoutput=1
\documentclass[11pt]{article}

\usepackage[final]{acl}
\usepackage{times}
\usepackage{latexsym}

\usepackage[T1]{fontenc}
\usepackage[utf8]{inputenc}

\usepackage{inconsolata}

\usepackage{graphicx}

\usepackage{multirow}
\usepackage{url}
\usepackage{booktabs}
\usepackage{linguex}
\usepackage{amsmath} 
\usepackage{tabularx}
\usepackage{enumitem}

\usepackage{todonotes}
\usepackage{comment}
\usepackage{subcaption}  % for subfigures
\usepackage{placeins}

\usepackage{listings}
\usepackage{footnote}
\usepackage{tablefootnote}
\usepackage{lipsum}
\usepackage{xcolor}
\usepackage{colortbl}
\usepackage{booktabs}

\usepackage{fontawesome5}

\title{References Matter: Investigating the Impact of Reference Set Variation\\ on Summarization Evaluation}

\author{
\textbf{Silvia Casola\thanks{\;Equal contribution.}\textsuperscript{\faMountain\kern1pt\faRobot}} \quad
 \textbf{Yang Janet Liu\footnotemark[1]\textsuperscript{\faBell}} \quad
 \textbf{Siyao Peng\footnotemark[1]\textsuperscript{\faMountain\kern1pt\faRobot}} \quad
 \\
 \textbf{Oliver Kraus\textsuperscript{\faMountain}} \quad
 \textbf{Albert Gatt\textsuperscript{\faSun}} \quad
 \textbf{Barbara Plank\textsuperscript{\faMountain\kern1pt\faRobot}}
\\
\textsuperscript{\faMountain} MaiNLP, 
LMU Munich
\quad
\textsuperscript{\faRobot} Munich Center for Machine Learning (MCML), Germany \\
\textsuperscript{\faBell} Department of Linguistics, University of Pittsburgh, USA \\
\textsuperscript{\faSun}  NLP Group, Department of Information and Computing Sciences, Utrecht University, Netherlands \\
\tt{
\{\href{mailto:s.casola@lmu.de}{\textcolor{black}{s.casola}},
\href{mailto:siyao.peng@lmu.de}{\textcolor{black}{siyao.peng}}, 
\href{mailto:o.kraus2@lmu.de}{\textcolor{black}{o.kraus2}}, 
\href{mailto:b.plank@lmu.de}{\textcolor{black}{b.plank}}\}@lmu.de}, \\
\tt{\href{mailto:jal787@pitt.edu}{\textcolor{black}{jal787@pitt.edu}}, 
\href{mailto:a.gatt@uu.nl}{\textcolor{black}{a.gatt@uu.nl}}
}}

\begin{document}
\maketitle
\begin{abstract}
Human language production exhibits remarkable richness and variation, reflecting diverse communication styles and intents. However, this variation is often overlooked in summarization evaluation.
While having multiple reference summaries is known to improve correlation with human judgments, the impact of the reference set on reference-based metrics has not been systematically investigated. 
This work examines the sensitivity of widely used reference-based metrics in relation to the choice of reference sets, analyzing three diverse multi-reference summarization datasets: SummEval, GUMSum, and DUC2004. We demonstrate that many popular metrics exhibit significant instability.
This instability is particularly concerning for n-gram-based metrics like ROUGE, where model rankings vary depending on the reference sets, undermining the reliability of model comparisons. 
We also collect human judgments on LLM outputs for genre-diverse data and examine their correlation with metrics to supplement existing findings beyond newswire summaries, finding weak-to-no correlation. Taken together, we recommend incorporating reference set variation into summarization evaluation to enhance consistency alongside correlation with human judgments, especially when evaluating LLMs.
\end{abstract}

\section{Introduction}
Human-written texts vary widely in terms of length, style, communicative intent, lexical/syntactical choices, and numerous other dimensions \cite{giulianelli-etal-2023-comes,liu-zeldes-2023-gumsum,PodcastSummary2022SIGIR,baan2023uncertaintynaturallanguagegeneration}. Such variation poses a significant challenge in the evaluation of summarization systems \cite{10.1007/s10579-017-9399-2, celikyilmaz2021evaluationtextgenerationsurvey}.
Traditional summarization metrics typically rely on comparing system outputs to one or more references, treating these references as a ``gold standard''. Although the limitations of  reference-based metrics have long been acknowledged \cite{rankel-etal-2013-decade,louis-nenkova2013,reiter-2018-structured, peyrard-2019-studying, fabbri-etal-2021-summeval, goyal2023newssummarizationevaluationera},
\begin{figure}[t]
  \centering
  \includegraphics[width=0.9\columnwidth]{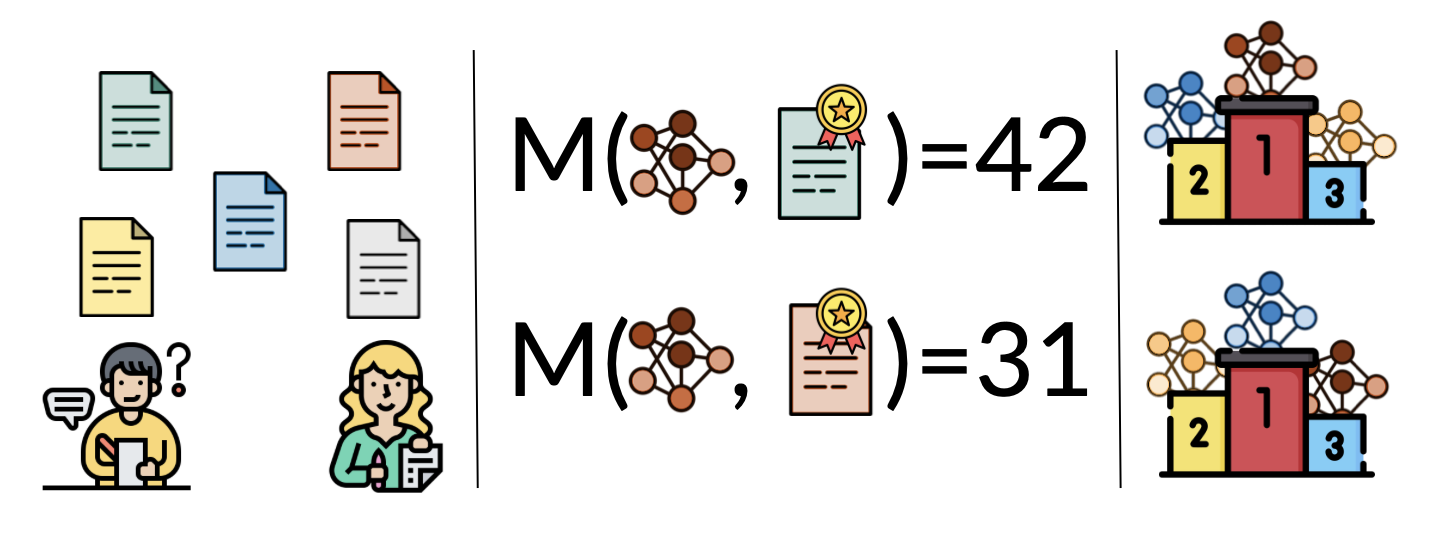}
  \caption{\textbf{Human-written summaries are diverse.} Using a human-written reference over another makes evaluation metrics fluctuate, and affects model ranking.}
  % \vspace{-10pt}
  \label{fig:figure1}
\end{figure}
\noindent
they remain widely popular due to their simplicity, low compute requirements, relative ease of adaptation to different languages, and reproducibility. 
%In 2024, out of $21$ ACL papers mentioning ``summarization'' in their title, $19$ ($90$\%) include a reference-based metric in their evaluation, with ROUGE \cite{lin-2004-rouge} being the most common ($71$\%), followed by BERTScore (\citealt{Zhang*2020BERTScore}, $52$\%).

The assumption behind the use of reference-based metrics is that system outputs that are more similar to the reference(s) are better, due to their ``human-likeness'' \cite{10.1613/jair.1.13715}. 
However, the significant variation in human-written summaries implies that evaluating system outputs against a single or limited set of references has inherent drawbacks.
Previous research has extensively looked at correlations between metrics and human judgments in summarization \cite{forde-etal-2024-evaluating, mondshine-etal-2025-beyond-n}, further exploring the use of multiple references to improve such correlations \cite{lin-2004-rouge, belz-reiter-2006-comparing, fabbri-etal-2021-summeval, tang-etal-2024-metrics} as well as the interpretability and efficiency aspects of such automatic metrics \cite{liu-etal-2023-towards-interpretable}. 
However, a much less studied question is \textbf{the extent to which automatic metrics are sensitive to the \emph{choice} of human-written reference summaries}, as shown in \autoref{fig:figure1}. In other words, are these metrics stable across different plausible gold-standard references? If metric scores vary significantly with the selected reference(s), this variation calls into question the reliability of many evaluation practices in the field.

In this work, we quantify the impact of reference choice on automatic evaluation metrics for summarization. 
Our contributions are as follows:
\begin{enumerate}
    \item \textbf{We investigate how different reference sets affect system rankings}.
    We show that system rankings based on n-gram-matching metrics (e.g., ROUGE) 
    strongly depend on the choice of the reference(s), undermining the reliability of model comparisons. However, rankings based on more semantically-oriented metrics exhibit greater stability. 
    \item \textbf{We examine the robustness of widely-used reference-based metrics for summarization at the instance and dataset level.} Our analysis reveals that the variation in scores introduced by the choice of reference on a dataset often exceeds the variation observed in state-of-the-art (SOTA) models. 
    \item \textbf{We collect new human judgment
    scores} on Large Language Model (LLM) outputs for the genre-diverse GUMSum \cite{liu-zeldes-2023-gumsum} dataset. We use these data \textbf{to reassess the correlation between automatic metrics and human judgments}, complementing earlier SummEval evaluations \citep{fabbri-etal-2021-summeval}, which were limited to pre-LLM models and newswire data. We find that correlations tend to increase with the number of references, and that the metric with the highest correlation varies depending on the evaluation dimension \emph{and} the number of references.
\end{enumerate}

\noindent Our analysis reveals that few metrics tend to show both reasonable correlation with human judgments \textit{and} robustness to the reference sets, especially when scoring LLM outputs.

The code is available at \url{https://github.com/mainlp/references-matter}.

\section{Related Work}
\label{sec:related-work}

\begin{comment}
    
\paragraph{Citations to add for INLG}
\begin{itemize}
\item \citet{zhang-etal-2024-benchmarking}: Benchmarking Large Language Models for News Summarization: 
``Despite major stylistic differences such as the amount of
paraphrasing, we find that LLM summaries
are judged to be on par with human written
summaries.''
\item GPT-Score \citep{fu-etal-2024-gptscore} and G-Eval  \citep{liu-etal-2023-g} both published on arxiv in Feb/Mar 2023 and later published on NAACL 2024  and EMNLP 2023
\item \citet{gao-etal-2025-llm-nlg-evaluation}: \url{https://direct.mit.edu/coli/article/51/2/661/128807/LLM-based-NLG-Evaluation-Current-Status-and}
\item \citet{liu-etal-2023-revisiting}: Revisiting the Gold Standard: Grounding Summarization Evaluation with Robust Human Evaluation (\url{https://aclanthology.org/2023.acl-long.228.pdf})
\item 
\citet{liu-etal-2024-learning}: On Learning to Summarize with Large Language Models as References
\item 
\citet{shen-etal-2023-large}: Large Language Models are Not Yet Human-Level Evaluators for Abstractive Summarization
\item multiref for MT: \citet{freitag-etal-2020-bleu}:  BLEU might be Guilty but References are not Innocent (\url{https://aclanthology.org/2020.emnlp-main.5.pdf}) and 
\citet{mathur-etal-2020-tangled}: 
Tangled up in BLEU: Reevaluating the Evaluation of Automatic Machine
Translation Evaluation Metrics (\url{https://aclanthology.org/2020.acl-main.448.pdf})
\item (maybe) \citet{pu2023summarizationalmostdead}: Summarization is (Almost) Dead
\item 
\end{itemize}

\end{comment}

\paragraph{Summarization Evaluation.}
Recent advances in Natural Language Generation (NLG) have significantly enhanced the development of summarization systems. However, their evaluation remains an open problem \cite{celikyilmaz2021evaluationtextgenerationsurvey, goyal2023newssummarizationevaluationera}.
Summarization evaluation metrics are broadly categorized into reference-based and reference-free \cite{10.1007/s10579-017-9399-2}. Reference-based metrics compare system outputs to human-written reference summaries, relying on methods such as n-gram overlap \cite{lin-2004-rouge, papineni-etal-2002-bleu}, embedding similarity \cite{ng-abrecht-2015-better, zhao-etal-2019-moverscore, Zhang*2020BERTScore}, or model-based  techniques \cite{peyrard-etal-2017-learning, scialom-etal-2019-answers, 10.5555/3540261.3542349}. 
In contrast, reference-free summarization metrics do not assume a gold standard \cite{10.5555/3540261.3542349, vasilyev-etal-2020-fill, gao-etal-2020-supert, gigant-etal-2024-mitigating}.  
More recently, growing research leverages LLMs as evaluators, with or without references 
\cite{song-etal-2024-finesure, li-etal-2024-leveraging-large}. 
% In the LLM-as-judge paradigm, prompting \cite{liu-etal-2023-g, bavaresco-etal-2025-llms} or the model's generative probability directly \cite{fu-etal-2024-gptscore} are typically used to evaluate the outputs.
In the LLM-as-judge paradigm, evaluations are typically based either on prompting the model to provide judgments \cite{liu-etal-2023-g, bavaresco-etal-2025-llms}, or on using its generative probabilities directly \cite{fu-etal-2024-gptscore}.

\paragraph{Metrics Meta-Evaluation.} 
Meta-evaluation of summarization metrics typically focuses on the extent to which they can be used as a proxy for human evaluation. \citet{reiter-belz-2009-investigation} examined the validity of automatic scores for NLG tasks, while \citet{rankel-etal-2013-decade} focused on ROUGE and its correlation with humans. \citet{peyrard-2019-studying} showed that metrics with reasonable correlation on lower-quality outputs tend to diverge when output quality increases. \citet{caglayan-etal-2020-curious} demonstrated the idiosyncrasies of automatic evaluation metrics, noting that high correlation with human judgments is not sufficient to characterize their reliability. \citet{fabbri-etal-2021-summeval} performed a large-scale meta-evaluation of summarization metrics, and found that most metrics have low correlation with human judgments on \textit{coherence}, while \textit{relevance} is 
weakly or moderately correlated. \citet{mondshine-etal-2025-beyond-n} performed a meta-analysis of reference-based, reference-free, and LLM-based summarization metrics focusing on eight languages from four typological families, showing low correlation. They also observed that off-the-shelf LLMs as judges still lag behind other metrics.

While most existing research focused on correlation to human scores, \citet{tang-etal-2024-metrics} addressed the challenge of evaluation when a limited number of references is available. They proposed leveraging LLMs to diversify the references, expanding the evaluation coverage and improving the correlation with humans. Their results show that increasing the number of references significantly enhances the reliability of existing evaluation metrics in terms of correlation. However, since LLM outputs tend to show less variability and follow distinct patterns compared to human-produced content \cite{giulianelli-etal-2023-comes, guo2024benchmarkinglinguisticdiversitylarge, shurofry2024growingtailincreasingoutput, doi:10.1073/pnas.2422455122}, relying on them to replace human references might introduce biases.
Beyond summarization, evaluation of reference variability has also been conducted in tasks such as machine translation \citep{castilho-2020-page, popovic-2021-agree, wu2025multiplereferencesmeaningfulvariations} and image captioning \citep{yi-etal-2020-improving}.

\begin{table*}[t!bh]
\centering
\resizebox{\textwidth}{!}{%
\begin{tabular}{c|cc|ccc|c|c}
\toprule
\multirow{2}{*}{dataset} & \multirow{2}{*}{\#doc} & \multirow{2}{*}{genre} & \multicolumn{3}{c}{references} & \multicolumn{2}{c}{outputs} \\ \cmidrule(l){4-6} \cmidrule(l){7-8} 
         &     &           & \#sums & \#chars & \#toks & model outputs & human judgments \\ \midrule
SummEval & 100 & news      & 1+10     & 226.3   & 43.1   & \citet{fabbri-etal-2021-summeval} + 4 LLMs & \citet{fabbri-etal-2021-summeval}  \\
GUMSum   & 48  & 12 genres & 5      & 291.3   & 52.1   & 4 LLMs & collected in this work  \\ 
DUC2004  & 489 & news      & 4      & 70.0    & 11.9   & 4 LLMs & n/a  \\
\bottomrule
\end{tabular}%
}
\caption{\textbf{Multi-reference summarization datasets.} \#sums indicates the number of human-written references per instance. We generate outputs using four LLMs and collect a new set of human judgments for GUMSum.}
\label{tab:datasets}
\end{table*}

\section{Experimental Setup}
\label{sec:experimental-setup}
To quantify the impact of human-written references on the scores of automatic metrics, we leverage multiple elements. For datasets, we use SummEval \cite{fabbri-etal-2021-summeval}, GUMSum \cite{liu-zeldes-2023-gumsum}, and DUC2004 \cite{duc2004}, which contain multiple human-written summaries (\S\ref{subsec:reference-summaries}), to assess how different reference summaries affect metric performances.
Next, to assess summarization models, we use the existing outputs provided by \citet{fabbri-etal-2021-summeval} for SummEval. As these outputs predate LLMs, we additionally collect outputs using LLMs (\S\ref{subsec:llm-summaries}) for all three datasets. Lastly, to compute the correlations with humans, we use the human judgments available in SummEval and gather new human ratings for GUMSum on both human and LLM-generated summaries (\S\ref{sub:human_judgments}). We prioritize GUMSum over DUC2004, as it includes multiple genres beyond news data. 
% We comply with the license of the existing datasets. For newly collected model outputs and human judgments, we follow the license of the corresponding underlying datasets. 
Our metric selection is outlined in \S\ref{subsec:metrics}.
Details on data licensing and codebase are provided in \autoref{sec:appendix_models}.

\subsection{Human-written Summaries}
\label{subsec:reference-summaries}

\textbf{SummEval} \cite{fabbri-etal-2021-summeval} is built on top of CNN/DM \cite{hermann2015teaching, nallapati-etal-2016-abstractive}, containing news articles and human-written highlights. The original authors selected $100$ instances from the test set and supplemented the existing highlights with ten additional reference summaries per instance, obtained via crowd-sourcing \cite{kryscinski-etal-2019-neural}.

\textbf{GUMSum} \cite{liu-zeldes-2023-gumsum} contains summaries created following general and genre-specific guidelines\footnote{\url{https://wiki.gucorpling.org/gum/summarization}} to function as a substitute for the source \cite{Nenkova2011McKeown}. 
We focus on the $48$ documents in the \texttt{dev} and \texttt{test} sets, which contain five human-written summaries each \cite{lin-zeldes-2024-GUMSAGE} across $12$ genres.

\textbf{DUC2004 Task1} \cite{duc2004} consists of $489$ news documents, most with four references. 
The guidelines allow the summaries to be in the form of short sentences or lists of keywords.\!\footnote{\url{https://duc.nist.gov/duc2004/tasks.html}} DUC2004 references are thus extremely concise (only up to $75$ characters).  The dataset has played a significant role in summarization research, being part of the annual TREC conference evaluation.

\autoref{tab:datasets} provides an overview of the three datasets. 
% While quality differences might exist at the annotator level, 
We treat all human references in the three datasets as ``gold'' since they were either authored by experts or validated through review.

\subsection{Model Outputs} 
\label{subsec:llm-summaries}

\citet{fabbri-etal-2021-summeval} collected model outputs for SummEval from $24$ extractive and abstractive summarization systems, which were SOTA between 2017 and 2019. We focus on the $16$ models for which they provided human judgments.

For all datasets, we also include summaries generated by contemporary LLMs. This is crucial given that prior studies demonstrated that evaluation metrics often show lower correlation with high-quality outputs \cite{peyrard-2019-studying, alva-manchego-etal-2021-un}. Below, we report a similar pattern for LLMs (\S\ref{subsec:corr_humans}).  
For consistency purposes, we follow \citet{lin-zeldes-2024-GUMSAGE} and use Llama3-3B-Instruct \cite{hermann2015teaching}, Qwen-2.5-7B-Instruct \cite{qwen2025qwen25technicalreport}, Claude-3.5 \cite{anthropic2024claude3.5}, and GPT-4o \cite{openai2024gpt4o}. 
For each LLM, we generate a single summary. We emphasize LLM variety over multiple generations.
Details on the generation parameters and prompts are reported in \autoref{sec:appendix_models}. 
% \todo{checked, ok}

\subsection{Human Judgments}
\label{sub:human_judgments}

SummEval \citep{fabbri-etal-2021-summeval} contains expert judgments 
that assess summaries based on four criteria: \textit{coherence}, \textit{consistency}, \textit{fluency}, and \textit{relevance}, using a Likert scale of $1$-$5$ \cite{zis-Likert1932A}.

To measure how well automatic metrics align with human judgments beyond the news domain, and to study whether findings on pre-LLM models align with those on LLM outputs, we collect a new set of human judgments using the same criteria on the $48$ GUMSum documents.
We hired three Master's students in Computational Linguistics and tasked them to evaluate four LLM outputs (\S\ref{subsec:llm-summaries}) and five human references (\S\ref{subsec:reference-summaries}), following \citet{fabbri-etal-2021-summeval}'s criteria.
\begin{table}[t]
\centering
\resizebox{\linewidth}{!}{
\begin{tabular}{c|cccc|cc} \toprule
Summarizer & Coh.$\uparrow$ & Con.$\uparrow$ & Flu.$\uparrow$ & Rel.$\uparrow$ & best$\uparrow$ & worst$\downarrow$ \\
\midrule
\texttt{claude} & \cellcolor{cyan!20} $\mathbf{4.75_{0.45}}$ & \cellcolor{cyan!20} $4.48_{0.65}$ & \cellcolor{cyan!20} $\mathbf{4.82_{0.42}}$ & \cellcolor{cyan!20} $\mathbf{4.26_{0.71}}$ & \cellcolor{cyan!20} $\mathbf{0.17_{0.38}}$ & \cellcolor{cyan!20} $\mathbf{0.03_{0.18}}$ \\ 
\texttt{gpt4o} & $4.42_{0.61}$ & \cellcolor{cyan!20} $\mathbf{4.61_{0.63}}$ & \cellcolor{cyan!20} $4.74_{0.45}$ & $4.07_{0.65}$ & \cellcolor{cyan!20} $0.11_{0.32}$ & $0.12_{0.32}$ \\ 
\texttt{Qwen} & \cellcolor{cyan!20} $4.66_{0.56}$ & $4.33_{0.84}$ & $4.68_{0.52}$ & \cellcolor{cyan!20} $4.23_{0.75}$ & \cellcolor{cyan!20} $0.16_{0.37}$ & $0.12_{0.33}$ \\ 
\texttt{Llama3} & \cellcolor{cyan!20} $4.71_{0.51}$ & $4.14_{0.97}$ & \cellcolor{cyan!20} $4.78_{0.42}$ & $4.09_{0.86}$ & \cellcolor{cyan!20} $0.12_{0.32}$ & $0.20_{0.40}$ \\ 
\midrule
humans & $4.54_{0.57}$ & $4.48_{0.71}$ & $4.70_{0.51}$ & $4.22_{0.69}$ & $0.09_{0.28}$ & $0.10_{0.31}$ \\ 
\bottomrule
\end{tabular}
}
\caption{{\textbf{Human judgments on GUMSum: LLM vs.~human-written summaries.} Above-human performances are highlighted in blue.}}
\label{tab:dimension_scores}
\end{table}
LLM-generated and human-written summaries were anonymized and shuffled. We also asked the evaluators to pick one best and one worst summary  for each document. 

\autoref{tab:dimension_scores} reports the results.  
Claude scored the best overall.
GPT-4o gets the highest \textit{consistency} but the lowest \textit{coherence} and \textit{relevance}, and is thus the least picked LLM. 
Interestingly, LLM outputs typically receive higher scores than human-written references. 
% This finding, in line with existing research \cite{zhang-etal-2024-benchmarking}, have potential important impact on reference-based evaluation as a whole, and opens questions on the use of (potentially lower quality) references for such high-quality outputs \cite{noh-etal-2024-beyond}. 
In line with previous work, e.g., \citet{zhang-etal-2024-benchmarking}, this finding has significant implications for reference-based evaluation and calls into question the use of potentially lower-quality references for assessing high-quality outputs \cite{noh-etal-2024-beyond}.
%While some previous research has proposed using LLM outputs as references given their perceived high quality, we refrain to do so in this paper since we focus on how \textit{human} production diversity impacts evaluation, while a growing pile of research is showing that, while high-quality, LLM-generated text has very distinct patterns with respect to humans \cite{}. \cite{}, for example, show that...

\subsection{Evaluation Metrics}
\label{subsec:metrics}
We consider several reference-based metrics, chosen to balance popularity and diversity. All metrics range in $0$--$100$. 
\autoref{sec:appendix_metrics} provides details. 

\textbf{ROUGE} \cite{lin-2004-rouge} is the most popular summarization metric. ROUGE-N computes n-gram overlap between a hypothesis and the references. ROUGE-L leverages the longest common subsequence, accounting for the word order. With multiple references, ROUGE considers the maximum or the mean of the n-gram overlap (ROUGE\textsubscript{max} and ROUGE\textsubscript{avg}). We report the F1-score. 

\textbf{BLEU} \cite{papineni-etal-2002-bleu}  is an n-gram metric primarily used for translation. It is precision-based and incorporates a brevity penalty. With multiple references, the n-gram count is clipped at the maximum count of n-grams in a single reference, and the length of the reference closest in size to the hypothesis is considered.

\textbf{METEOR} \cite{banerjee-lavie-2005-meteor} incorporates multiple linguistic aspects, including synonym matching, stemming, and word order, making it more robust in capturing semantic equivalence.
While primarily designed for translation, it has also been used to assess summaries. With multiple references, the maximum score is considered. 

\textbf{BERTScore} \cite{Zhang*2020BERTScore} leverages contextual embeddings and 
considers the cosine similarity between the embeddings of the hypothesis and the reference tokens. With multiple references, the final score is the maximum among the individual scores. We report the F1 score.

\textbf{BLEURT} \citep{sellam-etal-2020-bleurt}
is a model-based metric that leverages BERT fine-tuned on human judgments. The metric is not designed to handle multiple references; we compute scores for each reference and consider the maximum.

\begin{figure}[t]
\includegraphics[width=\columnwidth]{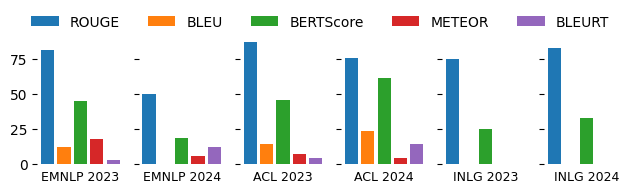}
  \caption{Percentage of papers about summarization that use common reference-based metrics.}
  % \vspace{-10pt}
  \label{fig:metric_use}
\end{figure}

To gain insights into recent metric usage, \autoref{fig:metric_use} summarizes the percentage of summarization papers from recent ACL, EMNLP, and INLG proceedings (detailed in \autoref{appendix:search}). % and their reported metric usage. %that report using each metric for evaluating model outputs at recent major NLP conferences. 
We found that reference-based metrics are still the most popular, with ROUGE in $79$\% of papers, followed by BERTScore ($44$\%). 
Our preliminary search also shows that LLM-as-a-judge evaluators are mainly (i.e., in $66$\% of the cases) used without references, placing them outside the scope of this paper.

\begin{figure*}[tb!]
    \centering
    \includegraphics[width=0.9\linewidth]{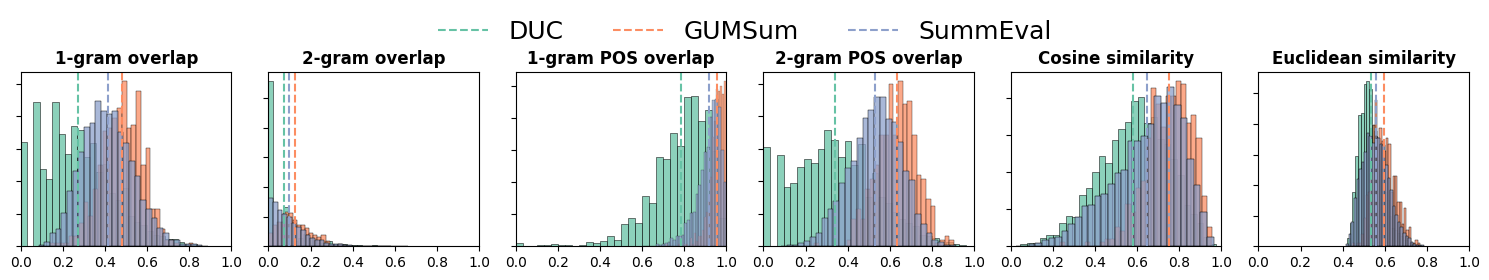}
    \caption{Variation in human-written summaries %at the lexical, syntactical and semantic level, 
    across datasets, measures inspired by~\citet{giulianelli-etal-2023-comes}.}
    \label{fig:probes_references}
\end{figure*}

\section{Reference Variability and Metric Robustness}
\label{sec:matrics-robustness}

Reference-based metrics assume that more human-like outputs deserve higher scores. However, human summaries are very diverse. This section examines how metrics fluctuate with different human references. 
By analyzing metric robustness, we aim to understand how conclusions about models, drawn from reference-based metrics, might change when different sets of human-written references are used, thereby undermining evaluation reliability.

\subsection{Human-written Summaries are Diverse} 
Human-written summaries show substantial diversity. 
We assess the variability in the multi-reference datasets 
following \citet{giulianelli-etal-2023-comes}.
For each pair of human-written summaries for the same instance, we report the lexical similarity (the overlapping distinct n-grams between two strings), the syntactic similarity (the overlap of part-of-speech tag n-grams), and the semantic similarity (the cosine and euclidean similarity between the embeddings of the two strings).

\autoref{fig:probes_references} shows these variations. At the dataset level, DUC and SummEval show the lowest similarity among human-written summaries across all dimensions. 
For GUMSum, summaries are more similar to each other. We hypothesize that this is likely due to the constrained annotation guidelines. It is also worth noting that the similarities revealed here are between different human-written summaries for a given instance as opposed to summaries across genres, for which we still expect significant variations, as demonstrated by \citet{liu-zeldes-2023-gumsum}. 
Overall, summaries tend to be similar at the syntactic level, less so at the semantic and lexical level. We also observe that LLM outputs show lower diversity (\autoref{appendix:llm_variability}), %\todo{checked, ok}
consistently with previous work \cite{giulianelli-etal-2023-comes}.

\subsection{Automatic Metrics Fluctuate Substantially at the Instance Level} 
\label{subsec:instance-level-var}

\begin{figure*}[b!th]
    \centering
    \includegraphics[width=0.85\linewidth]{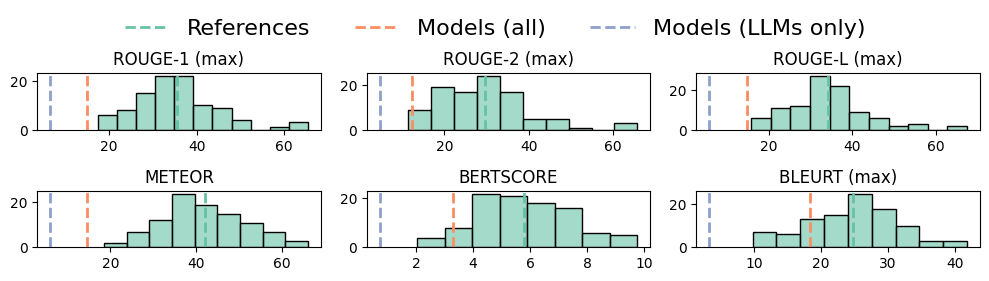}
    \caption{\textbf{Ranges of variability at the instance level on SummEval.} For each instance, we compute the range of the scores of the references against the remaining ones. The trends for ROUGE\textsubscript{max} and ROUGE\textsubscript{avg} are similar.}
    % \vspace{-10pt}
    \label{fig:instance_hist}
\end{figure*}

Given the diversity in human-written summaries, we quantify metric fluctuation at the instance level when using a different set of human-written references. For a metric $M$ and a set of human-written references $R = \{ r_1, r_2, \dots, r_N \} $, we compute $M(r_i, R-\{r_i\})$. Thus, for each document, we score each human-written summary using all the others as the reference set.
\begin{figure}[b!th]
    \centering    
    \includegraphics[height=0.6\textheight]{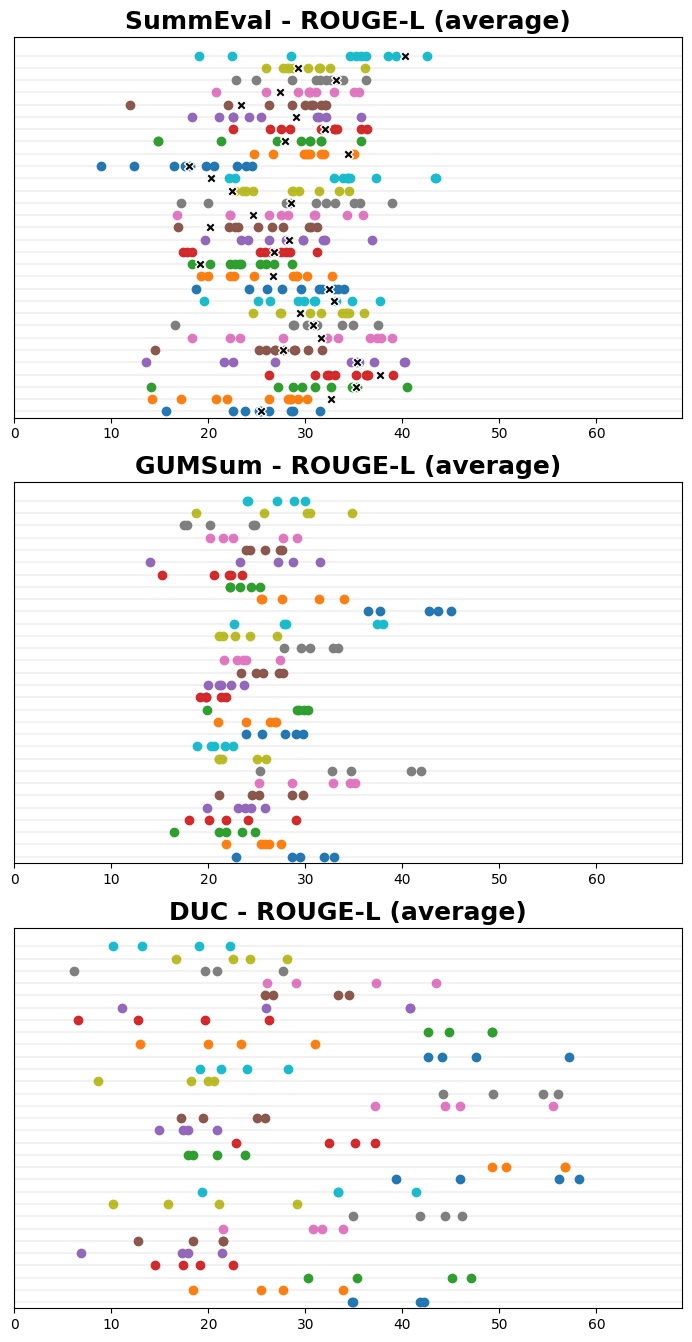}
    \caption{\textbf{Instance-level variation for ROUGE-L\textsubscript{avg}.} For every document (shown first $30$, one per line), we plot the score for every human-written reference against all other references (using the same color per source to aid interpretation). The original CNN/DM reference in SummEval is marked by a cross. 
    } 
    % \vspace{-12pt}
    \label{fig:rouge-l_instances}
\end{figure}
\autoref{fig:rouge-l_instances} exemplifies the observed instance-level variability 
%at the instance level
measured by ROUGE-L\textsubscript{avg} on the three datasets. 
For SummEval, we also mark the original reference (scraped highlights, \S\ref{subsec:reference-summaries}) in the CNN/DM dataset with a cross. The quality of these scraped references versus the ten later crowd-sourced ones is discussed further in \autoref{sec:appendix_cnndm}.

\paragraph{Scores assigned to human-written summaries are often low.} 
For example, the averaged ROUGE-L\textsubscript{avg} scores are $28.52_{\pm5}$, $27.46_{\pm3}$, $24.88_{\pm5.3}$ for SummEval, GUMSum, and DUC2004. 
Given the assumption that human reference summaries are of high quality, 
% (i.e., ``gold''),
metrics should produce high scores. Instead, they do not typically reflect this property.

\paragraph{Human-written reference scores vary widely.}
\autoref{fig:instance_hist} summarizes the instance-level variability of the individual scores (in \autoref{fig:rouge-l_instances}) for all evaluation metrics on SummEval (corresponding figures for GUMSum and DUC are in \autoref{appendix:hist}). For each metric, we compute the min-max range when scoring human-written references against all the others ($M(r_i, R-\{r_i\})$). \autoref{fig:instance_hist} shows the histogram of such ranges. 
%Note that 
\textbf{The ranges of variation observed within human-written references are, on average, very high}. 

Understanding the magnitude of such a range might not be obvious. For instance, an increase of $10$ points of BERTScore (typically scoring in the high range of the scale) might indicate a much larger improvement in performance than an increase of $10$ points of ROUGE-1.\!\footnote{The relative dynamics between metrics have been studied by \citet{kocmi-etal-2024-navigating} in machine translation.} To contextualize the magnitude of variation for each metric, we also report the performance range of summarization systems. Thus, for a model $S$, given its output $o_i$ for instance $i$, we score it through $M(o_i, R)$. Although these values are not directly comparable and should be interpreted with caution due to the use of different reference sets, they help contextualize the magnitude of the results and its potential impact on evaluation. For example, ROUGE-1\textsubscript{max} assigned to human-written references varies by about $35$ points on average (the green dashed line in \autoref{fig:instance_hist}), while the mean range is less than $20$ points across all model outputs (orange line), and much lower for LLM outputs (blue line). 
\textbf{Similarly, LLM summaries exhibit much less variability than human references on all metrics.}
% , despite different scales. 
These findings highlight the significance of 
% the observed 
variability and suggest that the ranking of summarization models is highly sensitive to the reference set.

\subsection{System Ranking Depends on the Reference(s) for N-Gram-Based Metrics}
\label{subsec:model-ranking-robustness}

\begin{figure*}[!t]
    \centering    
    \includegraphics[height=5cm]{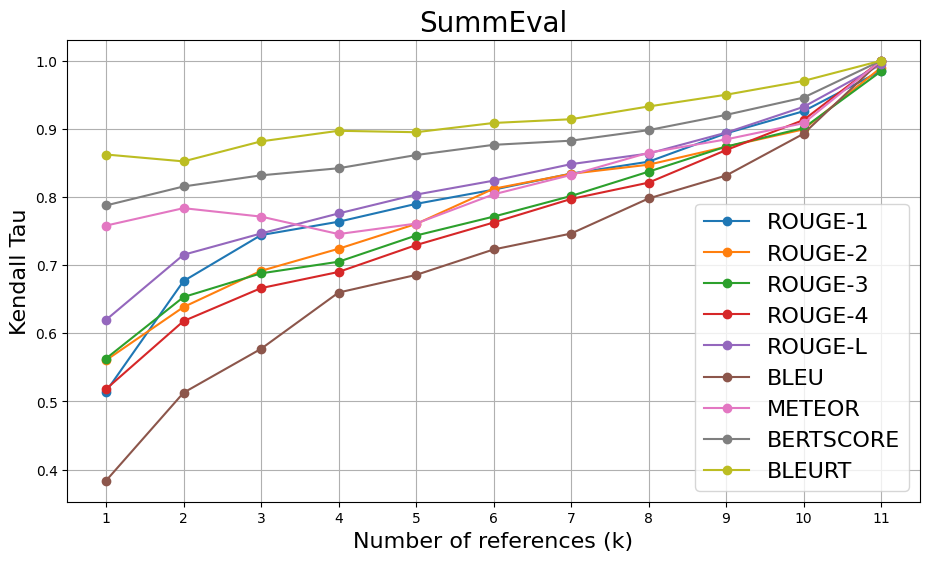}
    \includegraphics[height=5cm]{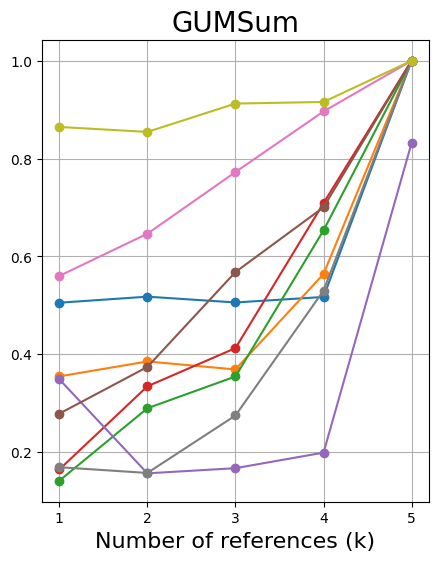}
    \includegraphics[height=5cm]{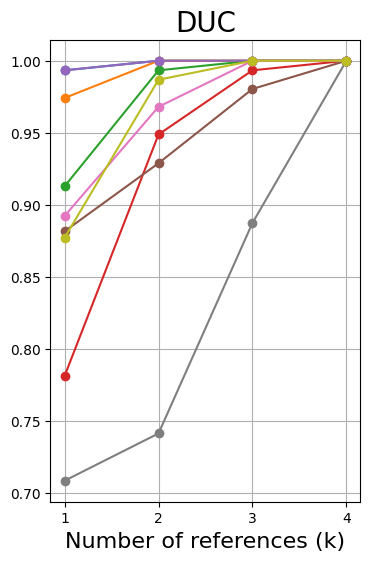}
    \caption{\textbf{Rank stability when increasing the number of references}. ROUGE\textsubscript{max} is presented. Note that we use different ranges for the \texttt{y} axis for each dataset to improve readability.}
    % \vspace{-10pt}
    \label{fig:rankings_N}
\end{figure*}

While we observed variability at the instance level, summarization metrics are typically designed to evaluate models across datasets, rather than individual instances. In this section, we investigate to what degree standard summarization metrics can handle the variability observed in human-written references when ranking summarization systems.

\paragraph{Procedure.} We sample $k$ human-written references ($k \in [1, N]$, where $N$ is the number of references for each document) from all available references for each instance. We then score the outputs of each summarization system using the same set of references.
Given $M$ systems $S_a$, $S_b$, \dots, $S_M$,  the metric induces a ranking $ S_a \succ S_b \succ \dots \succ S_M$. This process is repeated $100$ times, yielding $100$ rankings. 
We compute the pairwise Kendall rank correlation coefficient \cite{665905b2-6123-3642-832e-05dbc1f48979} between such ranks. 
High correlation indicates that models are similarly ordered, even when different sets of references are used.\footnote{Note that, for $k>1$, references in different reference sets might overlap, artificially increasing the observed correlation between rankings.}
% High correlation indicates that different sets of references lead to similar model ordering, even when different sets of references are used. 
\autoref{fig:rankings_N} reports the average correlation for pairs of ranks for each dataset and metric, from using $k$ human-written summaries as references. ROUGE\textsubscript{max} is shown in \autoref{fig:rankings_N}, and ROUGE\textsubscript{avg} is reported in \autoref{fig:rankings_N_mean} in \autoref{appendix:ranking}. 

\paragraph{Single Reference.} Evaluating with a single reference is common in summarization, as most datasets provide only one human-written summary.
\autoref{fig:rankings_N} (looking at $k=1$) shows the stability of different metrics with a single reference across the three datasets.
We find that \textbf{BLEU and ROUGE have very weak to moderate correlation between ranks across different references}.
In other words, using two different sets of plausible references would likely lead to different conclusions on relative model performance. 
We also notice a large variability among the individual pairs of rankings, with some showing negative correlation (refer to \autoref{tab:rank_1_all} in \autoref{appendix:ranking} for results on individual metrics and datasets). 

In contrast, \textbf{more semantically-oriented metrics show greater stability}. For SummEval, BLEURT shows the highest correlation between ranks, followed by METEOR and BERTScore. BLEURT and METEOR confirm their stability on GUMSum when ranking the LLM outputs. Other metrics (including BERTScore) show low or no correlation on GUMSum, with the exception of ROUGE-1. 
In all cases, metrics show much higher stability on DUC, for which all average correlations are above $0.7$. We speculate that high stability might stem from an artifact introduced by the short summary length required by the guidelines.

In summary, n-gram-matching metrics, though simple, are highly reference-dependent, undermining consistent model evaluation, while semantically-oriented ones show greater stability. Therefore, \textbf{we recommend always using model-based metrics in benchmarks with a single reference.} When cost is a factor, METEOR might offer a good balance of stability and affordability. % for quick evaluations.

\paragraph{Multiple References.} 
When scoring model outputs against a set of $k>1$ randomly sampled references, we observe that \textbf{the correlation between rankings obtained with different human-written references generally improves with an increased number of references}. This increased stability is expected and in line with similar findings that associate a larger number of references with a higher correlation with humans \cite{lin-2004-rouge}.

However, the \textbf{stability varies by metric}. ROUGE (especially ROUGE\textsubscript{max}) and BLEU tend to have low correlation between ranks. As an example, the ROUGE\textsubscript{max} scores require $5$-$10$ references to reach a level of stability that is comparable to that of BERTScore on SummEval with a single reference.  
ROUGE\textsubscript{avg} has a better stability than ROUGE\textsubscript{max}, especially with a larger set of references. For example, on SummEval, ROUGE-L\textsubscript{avg} has higher stability than BERTScore for $k>3$, while on GUMSum, ROUGE-2\textsubscript{avg} is the second most stable metric for $k>3$.
On all datasets, BLEURT and METEOR remain stable even with a single reference, with METEOR showing stability despite its simplicity. 

In general, trends on SummEval are clearer and simpler to interpret than the other two datasets. We speculate that this is due to the larger number of models used ($16$ pre-LLM models+$4$ LLMs on SummEval vs $4$ LLMs on GUMSum and DUC). BLEURT, METEOR, and BERTScore show the highest stability, while n-gram-based metrics show low to average correlation between ranks even when multiple references are used. 
The cases of GUMSum and DUC2004 are more complex to interpret and might be less meaningful given fewer model outputs (i.e., only four LLM outputs, which might increase the observed noise). For GUMSum, BLEURT continues to show high inter-rank correlation, with METEOR being the second most stable.
BERTScore, on the other hand, shows poor stability. Similar to the case with $k=1$, on DUC2004, all metrics show high stability, likely due to summaries being very short, as dictated by the guidelines.

\subsection{Correlation with Human Judgments}
\label{subsec:corr_humans}

\begin{figure*}[ht]
    \centering    
    \includegraphics[height=4cm]{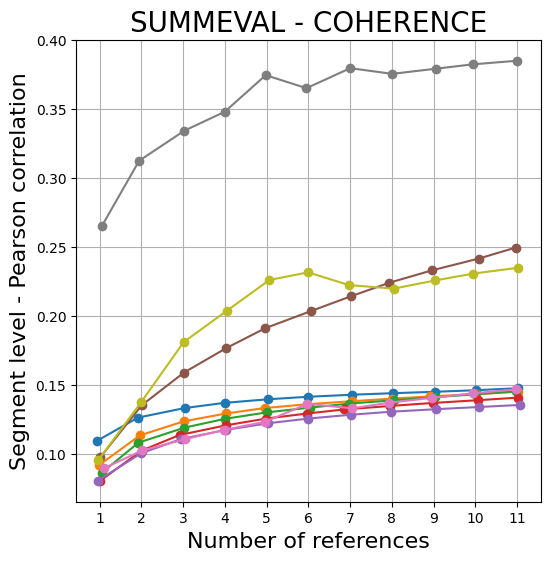}
    \includegraphics[height=4cm]{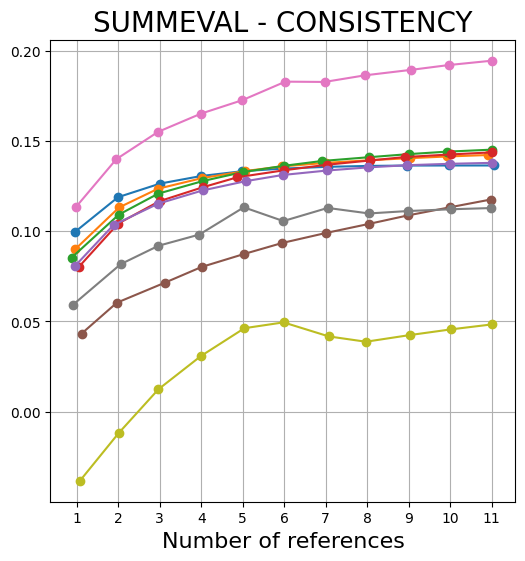}
    \includegraphics[height=4cm]{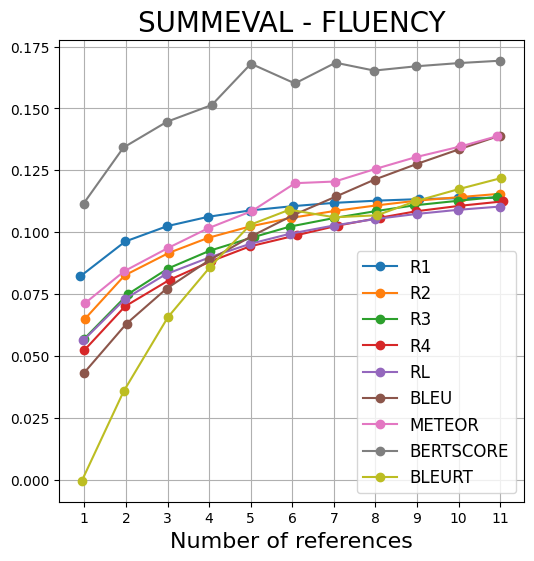}
    \includegraphics[height=4cm]{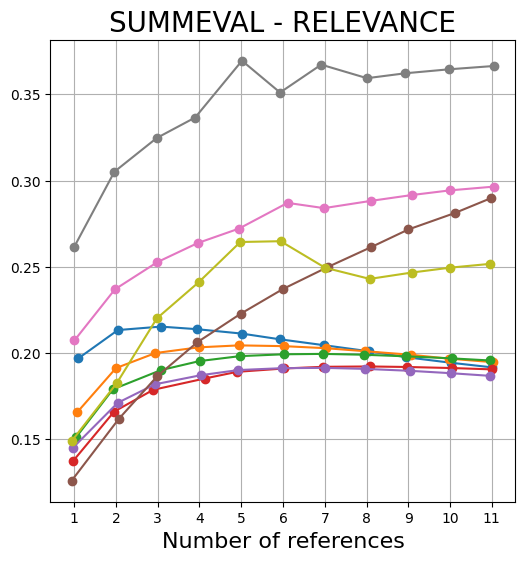}
    \includegraphics[height=4cm]{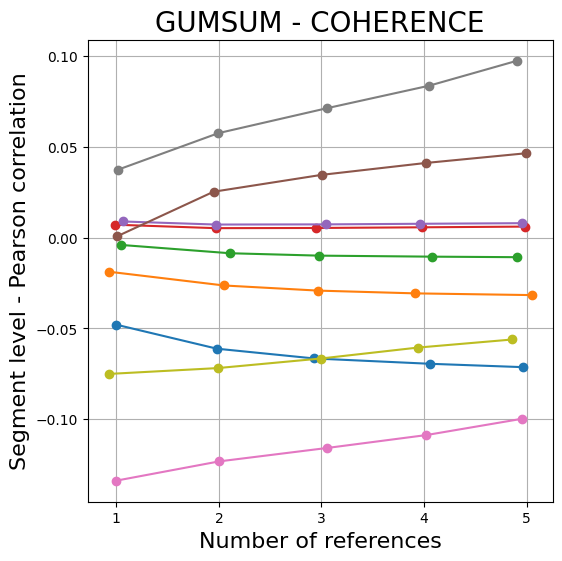}
    \includegraphics[height=4cm]{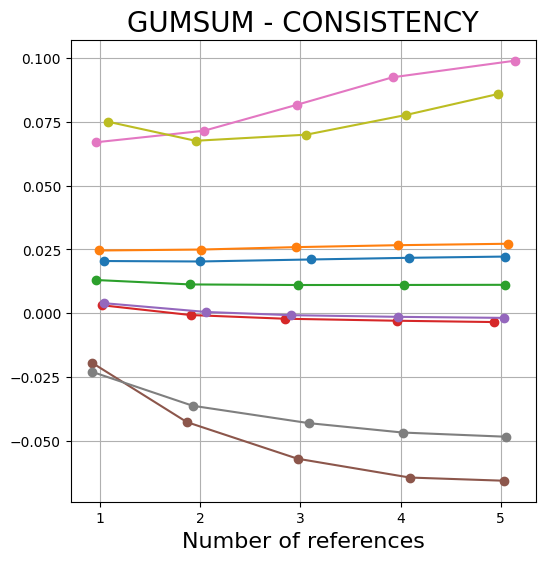}
    \includegraphics[height=4cm]{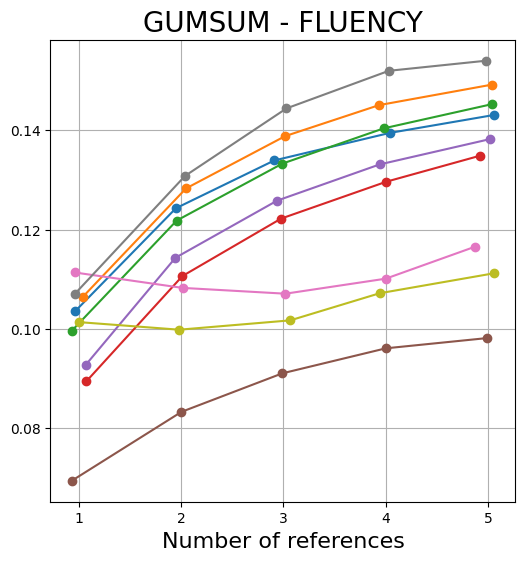}
    \includegraphics[height=4cm]{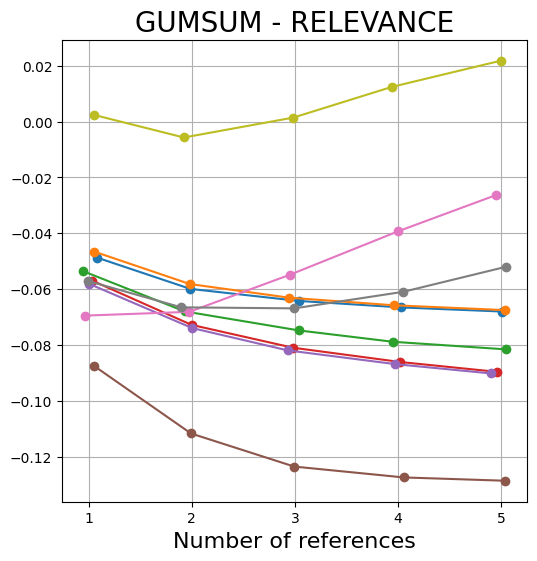}
    \caption{Pearson correlation at the instance level on SummEval (top) and GUMSum (bottom).}
    % \vspace{-10pt}
    \label{fig:corr_summ_instance}
\end{figure*}

In addition to stability, automatic metrics should correlate with humans. %, for which they are a proxy. 
We compute correlations for SummEval and GUMSum, for which we have human judgments,\!\footnote{For SummEval, we use the $16$ models studied by \citet{fabbri-etal-2021-summeval}; for GUMSum, the four LLMs.} at the instance and system level as the number of references $k$ increases.\!\footnote{For example, when considering two references, we consider the sets ($\binom{N}{2}$) of human-written references, where $N$ is the total number of references.
We compute the scores using such references as gold standard, and report the mean.}

\paragraph{Instance-level Correlation.}
\autoref{fig:corr_summ_instance} reports the instance-level correlation for SummEval (top) and GUMSum (bottom), respectively, versus the number of references. We show ROUGE\textsubscript{max}; corresponding figures using  ROUGE\textsubscript{avg} are in \autoref{appendix:sys-level-correlation}.

We notice \textbf{weak-to-no correlation} on both datasets. All correlations are generally higher on SummEval (where we consider outputs from the pre-LLM era) than on GUMSum (where we consider LLMs), in accordance to previous work showing that \textbf{correlation with human judgments decreases as the quality of the outputs improves} \cite{peyrard-2019-studying}. Additionally, reference-based evaluation itself could be problematic for very high-quality outputs when references are of worse quality than outputs (see Table \ref{tab:dimension_scores}, where model outputs are often scored on par with, or higher than, references) as also argued by \citet{goyal2023newssummarizationevaluationera}; metrics might also not be sensitive enough for outputs with more similar quality. The observed low correlation could be motivated by the low IAA in the GUMSum human judgments. 

For SummEval, increasing the number of references consistently leads to better correlation. This effect vanishes on GUMSum, where a larger reference set leads to no effect or slightly lower correlation. For SummEval, BERTScore shows the highest correlation on all dimensions but consistency, for which METEOR and ROUGE\textsubscript{avg} are better proxies. Notice how the best metric in terms of correlation with human judgment depends on the considered criterion \textit{and} the available number of references: BLEURT, for example, typically has low correlation when considering one reference only, performing worse than ROUGE. However, its performance improves when more references are considered, surpassing the scores of n-gram-based metrics.

\paragraph{System-level Correlation.}
System-level correlation is generally higher than instance-level correlation on SummEval; however, many criteria still show weak to moderate correlation when one or very few references are included. In most cases, such correlation tends to improve with the number of references. This is not the case for ROUGE\textsubscript{max}, especially when considering \textit{consistency}. 
The full results are provided in \autoref{fig:corr_summ_sys} in the \autoref{appendix:sys-level-correlation}. GUMSum is excluded from this analysis due to the small number of systems available.

\section{Conclusions}
\label{sec:conclusions}

In this work, we have investigated how reference sets impact the reliability of reference-based summarization metrics.
Our analysis across three multi-reference datasets reveals that, despite their popularity, token-matching metrics such as ROUGE are highly sensitive to the reference(s). This sensitivity leads to instability in system rankings, particularly when only a small number of references are available, which is typical in summarization datasets. In these situations, we thus recommend avoiding such metrics, echoing earlier calls for caution \cite{schmidtova-etal-2024-automatic-metrics}, 
in favor of model-based or reference-free alternatives.

We find that increasing the number of reference summaries consistently improves both the stability of metric scores and their alignment with human judgments. This might be explained by the possibility of representing a larger human diversity from the one hand, and from that of limiting annotator bias on the other. In these conditions, n-gram-based metrics such as ROUGE-N become more reliable. % and ROUGE\textsubscript{avg} should typically be preferred over ROUGE\textsubscript{max}.  
%However, when few references are available---especially in domains with high output variability---model-based metrics should be preferred, as they offer greater robustness despite their higher computational costs.

Our findings highlight the need to incorporate reference set variation into evaluation frameworks. Future metric development should explicitly account for this variation. 
While we have not specifically studied the dimensions of diversity in human references besides their lexical, syntactical, and semantic variation from \citet{giulianelli-etal-2023-comes}, we believe this is an important area of investigation. We speculate that the main challenge would be to collect a reference set that is ``diverse enough'' to represent human production for a fixed number of references. Future work in this direction needs to identify and characterize the relevant dimensions of diversity. These dimensions might be lexical, stylistic, intentional, or even sociolinguistic (cutting across the previously mentioned dimensions, \citealt{10.3389/frai.2024.1472411}). How to collect such references and whether specific guidelines should be adopted is also an open problem. Future work should also explore the role of genre.

We thus advocate for the creation of larger, more diverse multi-reference datasets, as well as for metric designs that are inherently robust to reference variability. Such efforts will be key to ensuring fairer and more reliable and human-aligned evaluation practices in summarization in the LLM era and beyond.

\section*{Limitations}

While our study highlights challenges posed by reference set variation in summarization evaluation, it also comes with several limitations. 

Although we focus on multi-reference datasets—SummEval, DUC2004, and GUMSum—such datasets remain relatively small. This reflects a broader limitation in current meta-evaluation practices, where multi-reference resources are the norm despite their limited scale.

Our analysis primarily targets standard evaluation criteria such as \textit{coherence}, \textit{consistency}, \textit{fluency}, and \textit{relevance}. While these are widely adopted and established, they do not capture all the nuances of summary quality, especially when the source texts are genre-diverse. More fine-grained human annotations and task-specific dimensions (e.g.,~factuality for news, stance for opinion pieces) would allow a deeper understanding of metric behavior under reference variation.

While we assess metric robustness using four systems (all of which are LLMs as opposed to the case for SummEval, where $16$ pre-LLM supervised systems are available) on the GUMSum and DUC2004 datasets, the relatively small number of models limits the generalization of the conclusions we can draw about system-level stability. Future work should include a larger and more diverse set of systems to better assess generalizability.

Moreover, using reference-based metrics for LLMs outputs (and generally, very high-quality outputs) has been questioned, especially when considering low-quality (e.g., scraped) references. While we focus on high-quality human-written references especially collected and checked for quality, we find that LLM-outputs are scored higher than references in our human evaluation campaign. We want to point out that the paper does not advocate to use reference-based metrics in such context; rather, it aims at shading lights on the limitation of the current evaluation practices, including the use of references in the cases in which the outputs might have higher quality, and its impact on stability and correlation. We also advocate for more research on the similarities and differences between LLM- and human-written summaries, to understand to which extent the use of LLM output as references could improve evaluation or rather introduce unwanted and largely unknown biases. 

Lastly, given our efforts to use contemporary LLMs, there remains a potential risk of data contamination. Since some of these datasets may have been seen during pretraining, this could affect both the outputs generated by LLMs and their evaluation scores. While we do not observe signs of memorization, we acknowledge that further controlled experiments are necessary to rigorously assess this risk.

\section*{Ethics Statement}
All annotators involved in the collection of human judgments for the GUMSum dataset have given their consent to participate in the study and to allow us to publish the collected scores. They were paid according to national standards.

We used AI-assistants for improving the clarity of the text and of the figures.

\section*{Acknowledgments}
\label{sec:acknowledgments}

We would like to thank the members of the MaiNLP lab for their valuable feedback, especially to Andreas Säuberli, Diego Frassinelli, Shijia Zhou, Soh-Eun (Ryan) Shim, and Domenico De Cristofaro.
We also thank Pingjun Hong, Yuchen Mao, and Sebastian Loftus for their annotations. 
We recognize the support for Yang Janet Liu\footnote{Work was carried out while at MaiNLP, LMU Munich.} and Barbara Plank through the ERC Consolidator Grant DIALECT 101043235.

\bibliography{custom}

\clearpage

\appendix
\section{LLM-generated Summaries}
\label{sec:appendix_models}
For each source, we generate a single summary using each of the four LLMs. For SummEval, the sources corresponding to the $100$ multi-reference summaries are taken into account. For DUC2004 Task 1, we generate summaries for the whole dataset ($489$ instances). For GUMSum, we focus on the dev and test set (in total $48$ instances). Thus, we generate a total of $2548$ summaries. 
We comply with the license of the existing datasets. For newly collected model outputs and human judgments, we follow the license of the corresponding underlying datasets. The codebase will be made publicly available upon publication.

\subsection{Prompts}
\subsubsection{SummEval}

\texttt{
Article: \{full\_text\}. Summarize the article in three sentences. Summary:
}

\subsubsection{DUC2004 Task 1}

\texttt{
The task is to create a very short single-document summary for the article below. }

\texttt{A very short summary should not be longer than 75 characters - this includes spaces and punctuation. }

\texttt{We will chop off characters beyond the 75th, so please do not include more than 75. }
    
\texttt{A very short summary could look like a newspaper headline, be a list of important terms or phrases separated by commas, a sentence, etc. }

\texttt{It should not contain any formatting, i.e., no indented lists, etc. Feel free to use your own words.}
    
\texttt{Article: \{full\_text\}}
\texttt{Summary:}

\subsubsection{GUMSum}
Following \citet{liu-zeldes-2023-gumsum}, a general prompt was used to instruct LLMs to generate summaries, as shown below. 
\texttt{Summarize the following article in 1 sentence. Make sure your summary is one sentence long and does not exceed 380 characters. Example of summary style: {example}}

\texttt{\{doc\_text\}}

\texttt{Summary:}

\subsection{LLM Output Evaluation}

\autoref{tab:llms} reports the scores obtained by the four LLMs when using all available references for scoring. 

\begin{table*}[t]
\centering
\resizebox{1.0\linewidth}{!}{
\begin{tabular}{l|c c c c c c c | c c c c c c c | c c c c c c c}
\toprule
 & \multicolumn{7}{c|}{\textbf{SummEval}} & \multicolumn{7}{c|}{\textbf{DUC}} & \multicolumn{7}{c}{\textbf{GUMSum}} \\
 & R-1 & R-2 & R-L & BLEU & MTR & BS & BLRT & R-1 & R-2 & R-L & BLEU & MTR & BS & BLRT & R-1 & R-2 & R-L & BLEU & MTR & BS & BLRT \\
\midrule
Qwen & 36.43 & 15.71 & 31.56 & 17.41 & 42.07 & 89.67 & 55.15 & 42.06 & 17.13 & 36.67 & 10.03 &  31.98&  89.96&  49.28&  44.01& 19.23&  34.20&  22.90&  35.25&  90.23&  47.87\\
Llama3 & 37.50 & 18.09 & 32.79 & 22.07 & 46.09 & 90.04 & 56.61 & 37.44 & 12.74 & 32.79 & 6.45 &  24.97&  89.87&  46.01&  43.51& 19.96&  35.34&  24.65&  36.70&  90.33&  48.50\\
Claude & 36.03 & 17.80 & 31.89 & 20.84 & 46.75 & 89.83 & 56.07 & 47.69 & 20.92 & 41.73 & 12.86 &  39.48&  91.27&  56.94&  44.99& 18.65&  34.51&  20.85&  37.62&  90.40&  50.13\\
GPT4 & 35.86 & 17.90 & 31.80 & 21.84 & 46.36 & 90.10 & 57.22 & 45.73 & 19.22 & 39.27 & 12.07 &  38.12&  91.15&  56.01&  47.02& 19.12&  34.23&  21.33&  43.46&  90.17&  51.61\\
\bottomrule
\end{tabular}
}
\caption{\textbf{LLM scores on the multi-reference datasets.} \textit{R}, \textit{MTR}, \textit{BS}, and \textit{BLRT} are short for ROUGE, METEOR, BERTScore, and BLEURT respectively.  All available references are used in the evaluation. For ROUGE and BLEURT, we consider the max-variation of the score.}
\label{tab:llms}
\end{table*}

\section{Reference-based Metrics}
\label{sec:appendix_metrics}
\paragraph{ROUGE.}
We use the \texttt{sacrerouge}\footnote{\url{https://github.com/danieldeutsch/sacrerouge}} python implementation of ROUGE \cite{deutsch-roth-2020-sacrerouge}, with default parameters. 

\begin{lstlisting}[language=Python]
    rouge = Rouge(
    max_ngram = 4,
    use_porter_stemmer = True,
    remove_stopwords = False,
    max_bytes = None,
    max_words = None,
    compute_rouge_l = True,
    skip_bigram_gap_length = None,
    scoring_function = "max", # or "average"
)
\end{lstlisting}
Notice that the implementation of ROUGE uses the Jackknife method when multiple references are provided.

\paragraph{BLEU.}
We use the \texttt{sacrebleu}\footnote{\url{https://github.com/mjpost/sacrebleu}} python implementation of BLEU \cite{post-2018-call}, with default parameters. 

\begin{lstlisting}[language=Python]
    bleu = BLEU(
    lowercase=False, 
    force=False, 
    tokenize=tokenize,
    smooth_method='exp', 
    smooth_value=None,
    effective_order=False) # True when used at the sentence level
)
\end{lstlisting}

\paragraph{METEOR.}
We use the Hugging Face version of Meteor, implemented through the \texttt{evaluate}\footnote{\url{https://github.com/huggingface/evaluate}} library, with default parameters, which wraps the NLTK implementation of the metric.\!\footnote{\url{https://www.nltk.org/api/nltk.translate.meteor_score.html}} 

\begin{lstlisting}[language=Python]
nltk.translate.meteor_score.meteor_score(references: ~typing.Iterable[~typing.Iterable[str]], hypothesis: ~typing.Iterable[str], preprocess: ~typing.Callable[[str], str] = <method 'lower' of 'str' objects>, stemmer: ~nltk.stem.api.StemmerI = <PorterStemmer>, wordnet: ~nltk.corpus.reader.wordnet.WordNetCorpusReader = <WordNetCorpusReader in '/Users/stevenbird/nltk_data/corpora/wordnet'>, alpha: float = 0.9, beta: float = 3.0, gamma: float = 0.5) -> float[source]

\end{lstlisting}

\paragraph{BERTScore.}
We use the Hugging Face version of BERTScore, implemented through the   \texttt{evaluate}\footnote{\url{https://github.com/huggingface/evaluate}} library, with default parameters, which wraps the \texttt{bert\_score} implementation.\!\footnote{\url{https://github.com/Tiiiger/bert_score}} No TF-IDF weighting is used.
Embeddings are obtained by using \texttt{FacebookAI/roberta-large}. We did not fine-tune the model. The corresponding hash is \texttt{roberta-large\_L17\_no-idf\_version=0.3.12 \\ (hug\_trans=4.48.3)}.

\paragraph{BLEURT.}
We use the original Google version of BLEURT.\!\footnote{\url{https://github.com/google-research/bleurt}} We used the recommended checkpoint,\!\footnote{available at \url{https://storage.googleapis.com/bleurt-oss-21/BLEURT-20.zip}} which we did not fine-tune. 

\section{Use of Reference-based Metrics}
\label{appendix:search}
We survey papers published at ACL, EMNLP, and INLG from 2023 and 2024. Given the list of long and short accepted papers, we collected papers matching the keyword \texttt{summar*} in their title. After filtering out papers that are not about summarization research (e.g., \textit{summarizing} the state of the art for a different topic), for each paper we checked whether one of our chosen metrics had been used for evaluation. Finally, we point out that some of the surveyed papers do not perform experimental work and thus do not evaluate model outputs (e.g., paper studying human evaluation); the reported percentages are thus slightly underestimated.

\section{Variability in Humans and LLMs}
\label{appendix:llm_variability}

\begin{figure*}[h!bt]
    \centering
    \begin{subfigure}[b]{\linewidth}
    \centering
    \includegraphics[width=\linewidth]{figures/probes_on_ref.png}
    \caption{human-written summaries}
    \end{subfigure}
     \begin{subfigure}[b]{\linewidth}
    \centering
    \includegraphics[width=\linewidth]{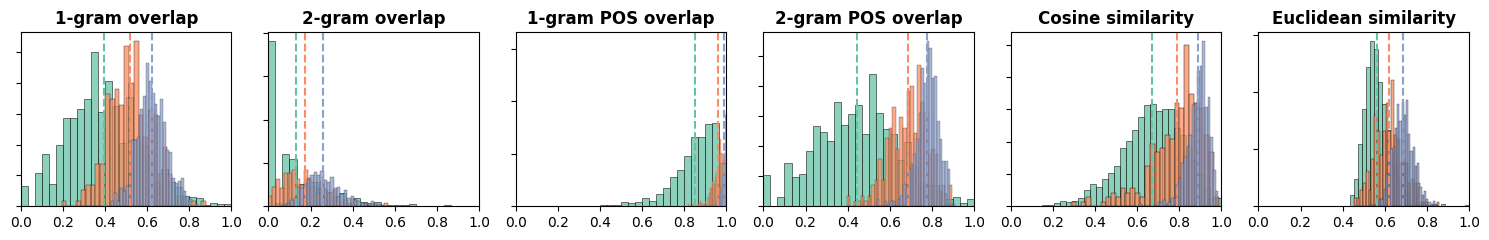}
    \caption{LLM-generated summaries}
    \end{subfigure}
    \caption{Variation in human-produced summaries (top) and LLM-generated summaries (bottom).}
    \label{fig:probes_references_appx}
\end{figure*}

\autoref{fig:probes_references_appx} compares the variability observed in human-written summaries and in LLM-generated ones.
To characterize LLM-generated summaries, given and instance $i$ and a pair of human-written summaries for instance $i$ $R_{ji} $ and $R_{zi}$, we plot $P(o_i, o_j)$ where $P$ is the lexical, syntactic, or semantic similarity.
To characterize LLM-generated summaries, given and instance $i$  system $S_j$ and $S_z$, producing outputs $o_j$ and $o_z$, we plot $P(o_{ji}, o_{zi})$. 

When compared to summaries generated by different humans, those produced by various LLMs exhibit far less variation, particularly at the semantic and syntactic levels. Quantifying and mitigating the reduced richness and variability of LLM-generated content is an area of ongoing research \cite{guo2024benchmarkinglinguisticdiversitylarge, shurofry2024growingtailincreasingoutput, giulianelli-etal-2023-comes} and, while not reflecting poor output quality, raises an open question about whether LLMs can fully replace human references.

\section{Quality of Scraped References}
\label{sec:appendix_cnndm}
\begin{figure*}[t!bh]
    \centering
    \includegraphics[width=\linewidth]{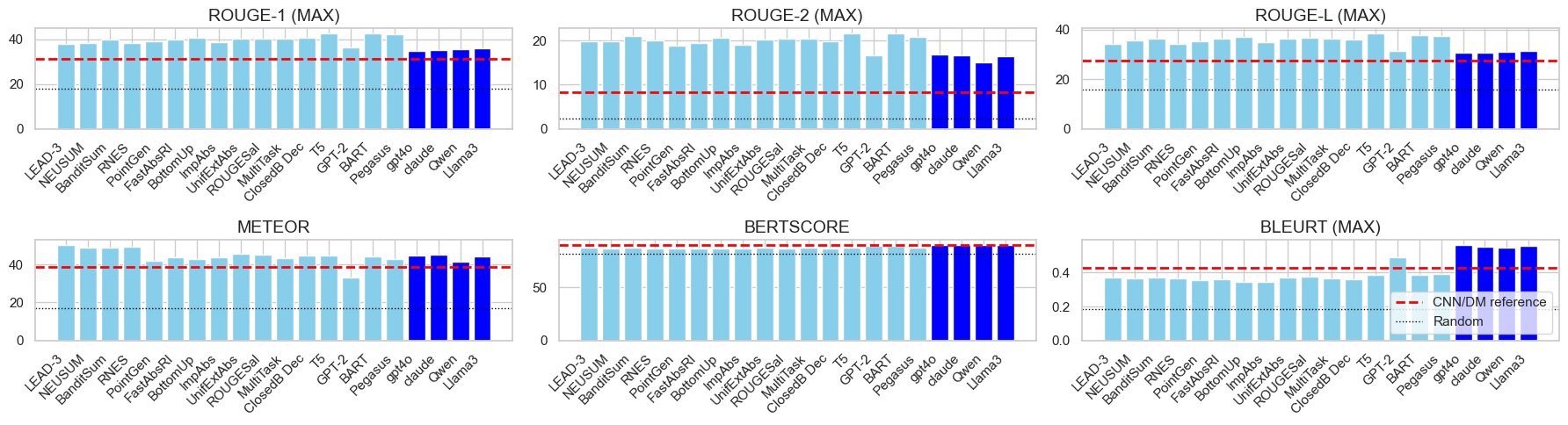}
    \caption{\textbf{Models and CNN/DM reference compared to $10$ crowd-sourced references in SummEval.} SummEval models are in lighter blue, LLMs in darker blue. The red line indicates the CNN/DM reference score, and the black dotted line represents a random baseline using a different news article as the hypothesis.}
    \label{fig:old_vs_new}
\end{figure*}

One peculiarity of summarization 
datasets is that they are typically gathered from existing sources \cite{dahan-stanovsky-2025-state}. 
This is the case of the CNN/DM dataset \cite{hermann2015teaching, nallapati-etal-2016-abstractive}, where articles are scraped from news websites, with the concatenated highlights acting as the target summary. The quality of such targets has been criticized in previous studies \cite{kryscinski-etal-2019-neural, srivastava-etal-2023-hiding}, as they contain extraneous facts, references to other articles, and other issues.

To better investigate the impact of these issues---and thus, quantify the gap between scraped summaries and crowd sourced ones---we leverage the additional references in SummEval. To this end, we treat the scraped CNN/DM references as hypotheses, and we score them against the 10 crowd-sourced ones in the  SummEval dataset. We assume the latter provide a superior reference set for two main reasons: a) crowd-sourced references are specifically collected to act as summaries, with clear guidelines and an emphasis on quality; b) the larger cardinality of the crowd sourced human references (cf.\ \autoref{tab:datasets}) allows for a more comprehensive view of human-produced summaries. We compare these score to those of system outputs.

\autoref{fig:old_vs_new} reports the scores of the original CNN/DM references (red line) and those of the outputs of the systems (blue bars). We also report a random baseline (dotted line) in which a summary of a different document randomly sampled from the collection is used as hypothesis.

Notably, the original CNN/DM references perform \emph{worse} than all model outputs in all cases but one when evaluated using n-gram-matching metrics. 
When using BERTScore, CNN/DM receives a score close to that of the outputs from LLMs.  
BLEURT rates the original reference higher than SummEval system outputs (except for GPT-2), but still lower than all LLM-generated outputs. These observations corroborate previous concerns on reference summary quality, especially when used to score high-quality outputs \cite{goyal2023newssummarizationevaluationera}.

\section{Instance-level variation}
\label{appendix:hist}

\begin{figure*}[t!bh]
    \centering
    \includegraphics[width=0.9\linewidth]{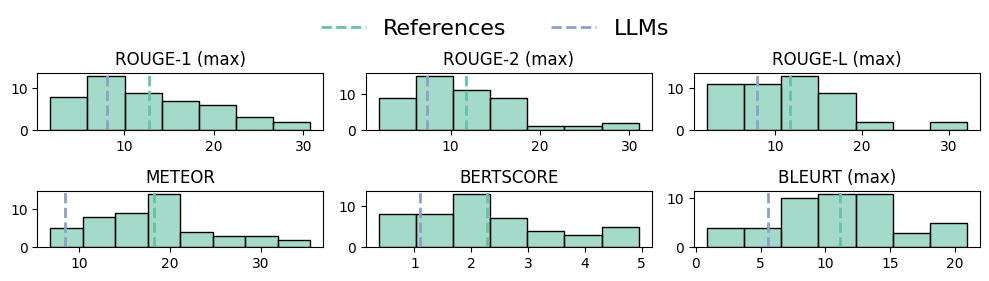}
    \caption{\textbf{Ranges of variability at the instance level on GUMSum.} For each instance, we compute the range of the scores of the references scored against the remaining ones.}
    \label{fig:instance_hist_gum}
\end{figure*}

\begin{figure*}[t!bh]
    \centering
    \includegraphics[width=0.9\linewidth]{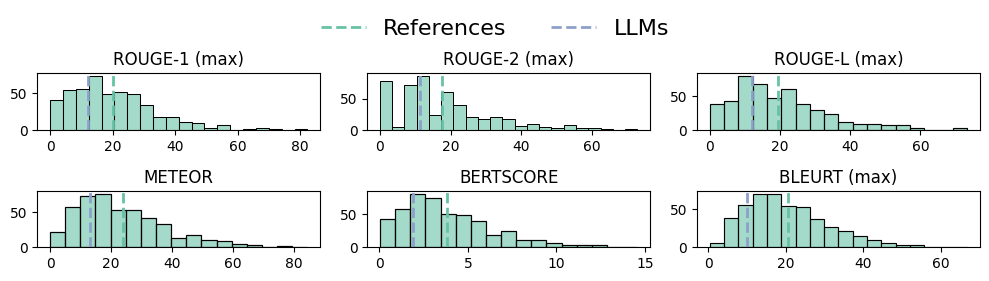}
    \caption{\textbf{Ranges of variability at the instance level on DUC.} For each instance, we compute the range of the scores of the references scored against the remaining ones.}
    \label{fig:instance_hist_duc}
\end{figure*}

\autoref{fig:instance_hist_gum} and \autoref{fig:instance_hist_duc} contain the histogram of the instance-level variability for GUMSum and DUC respectively. 


\section{Model Ranking}
\label{appendix:ranking} 

\autoref{tab:rank_1_all} contains the analysis of the rank stability when using one single reference ($k=1$). For SummEval, we considering the cases of ranking the models studied by \citet{fabbri-etal-2021-summeval}, the LLMs and the combination of the two separately.

\begin{figure*}[!t]
    \centering    
    \includegraphics[height=5.1cm]{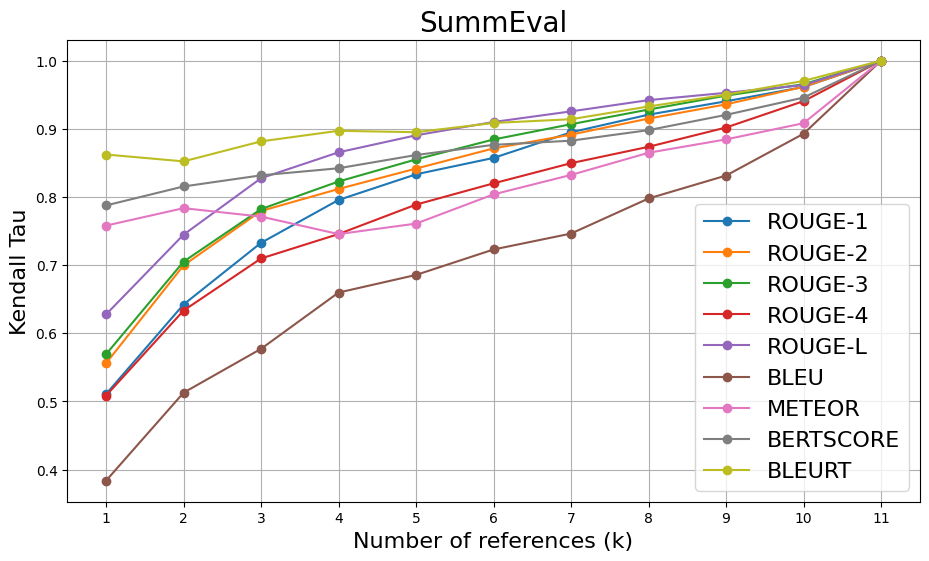}
    \includegraphics[height=5.1cm]{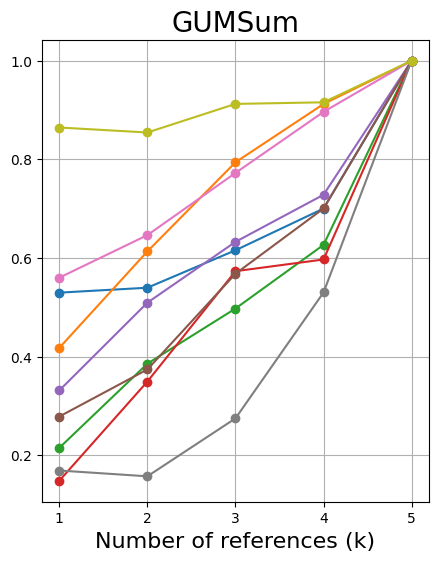}
    \includegraphics[height=5.1cm]{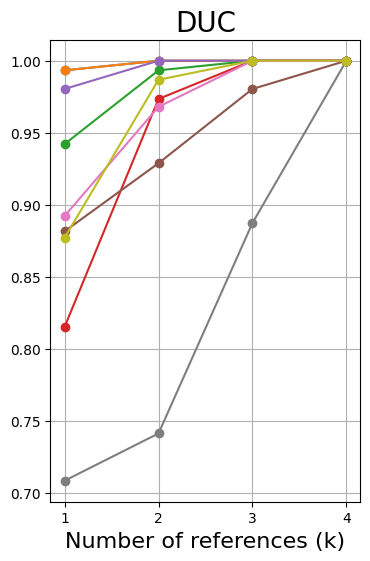}
    \caption{\textbf{Rank stability when increasing the number of references} over all three datasets. For ROUGE, we show the mean variant. Notice we use different ranges for the y axes for each dataset to improve readability. }
    \label{fig:rankings_N_mean}
\end{figure*}

\autoref{fig:rankings_N_mean} shows the Kendall tau correlation between ranks as the number $k$ of references increases. We use the mean version of ROUGE.

\begin{table*}[h!]
\resizebox{\linewidth}{!}{%
\begin{tabular}{cc|cccccccccccccc}
\toprule
&  & R1\textsubscript{mean} & R2\textsubscript{mean} & R3\textsubscript{mean} & R4\textsubscript{mean} & RL\textsubscript{mean} & R1\textsubscript{max} & R2\textsubscript{max} & R3\textsubscript{max} & R4\textsubscript{max} & RL\textsubscript{max} & BLEU & MTR & BS & BLRT \\
\hline
\multirow{4}{*}{SummEval\textsubscript{all models}} & min & -.04 & -.07 & .05 & -.09 & .12 & -.06 & -.09 & -.04 & -.06 & .04 & -.27 & .41 & .54 & .67 \\
& avg & .51 & .56 & .57 & .51 & .63 & .51 & .56 & .56 & .52 & .62 & .38 & .76 & .79 & .86 \\
& std & .16 & .14 & .12 & .14 & .12 & .18 & .15 & .13 & .13 & .14 & .16 & .07 & .06 & .04  \\
& max & .91 & .87 & .91 & .82 & .92 & .87 & .86 & .88 & .85 & .92 & .82 & .94 & .95 & .98 \\
\hline
\multirow{4}{*}{SummEval\textsubscript{pre-LLMs}} & min & -.05 & -.03  & -.05 & -.15 & 0 & -.05 & -.03 & -.12 & -.28 & .-07 & -.30 & .57 & .28 & .50 \\
& avg & .48 & .49 & .48 & .39 & .55 & .49 & .50 & .57 & .39 & .55 & .39 & .78 & .68 & .79 \\
& std & .14 & .13 & .13 & .15 & .13 & .13 & .13 & .15 & .16 & .12 & .17 & .06 & .09 & .07  \\
& max & .87 & .88 & .90 & .83  & .90 & .88 & .85 & .87 & .88 & .90 & .85 & .97 & .93 & .97 \\
\hline
\multirow{4}{*}{SummEval\textsubscript{LLMs}} & min & -.33 & 0 & -.33 & -1 & 0 & -.33 & -.33 & .67 & -1 & 0 & -.67 & 0 & -67 & 0  \\
& avg & .59 & .54 & .49 & .48 & .59 & .58 & .52 & .48 & .49 & .57 & .59 & .60 & .62 & .80 \\
& std & .31 & .32 & .33 & .35 & .31 & .32 & .32 & .34 & .36 & .31 & .32 & .31 & .28 & .21  \\
& max & 1 & 1 & 1 & 1 & 1 & 1 & 1 & 1 & 1 & 1 & 1 & 1 & 1 & 1\\

\hline
\multirow{4}{*}{GUMSum}  & min & 0 & -1 & -1 & -1 & -1 & 0 & -1 & -1 & -1 & -1 & -1 & 0 & -1 & .33 \\
& avg  & .53 & .42 & .22 & .15 & .33 & .51 & .35 & .14 & .16 & .35 & .28 & .56 & .17 &.86 \\
& std & .32 & .40 & .44 & .47 & .41 & .32 & .43 & .47 & .47 & .43 & .45 & .31 & .48 & .19  \\
& max & 1 & 1 & 1 & 1 & 1 & 1 & 1 & 1 & 1 & 1 & 1 & 1 & 1 & 1  \\
\hline
\multirow{4}{*}{DUC} & min & .67 & .67 & .33 & -.33 & .67 & .67 & .67 & .33 & -.33 & .67 & .33 & .67 & .33 & .67 \\
& avg  & .99 & .99 & .94 & .82 & .98 & .99 & .97 & .91 & .78 & .99 & .88 & .89 & .71 & .88 \\
& std & .05 & .05 & .12 & .22 & .08 & .05 & 0.9 & .15 & .25 & .05 & .17 & .16 & .24 & .16 \\
& max & 1 & 1 & 1 & 1 & 1 & 1 & 1 & 1 & 1 & 1 & 1 & 1 & 1 & 1 \\ \bottomrule 
\end{tabular}%
}
\caption{\textbf{Rank stability with a single reference.} We ranked systems $100$ times and compute the Kendall tau correlation among such rankings. \textit{R}, \textit{MTR}, \textit{BS}, and \textit{BLRT} are short for ROUGE, METEOR, BERTScore, and BLEURT respectively.}
\label{tab:rank_1_all}
\end{table*}

\section{System-level Correlation}
\label{appendix:sys-level-correlation}
\begin{figure*}[t!bh]
    \centering    
    \includegraphics[height=4cm]{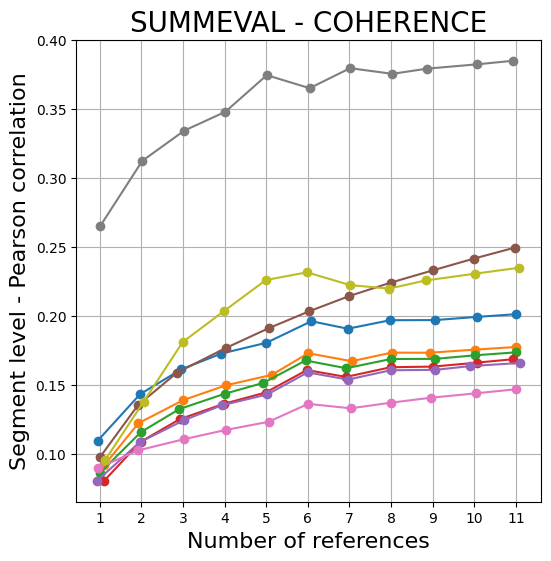}
    \includegraphics[height=4cm]{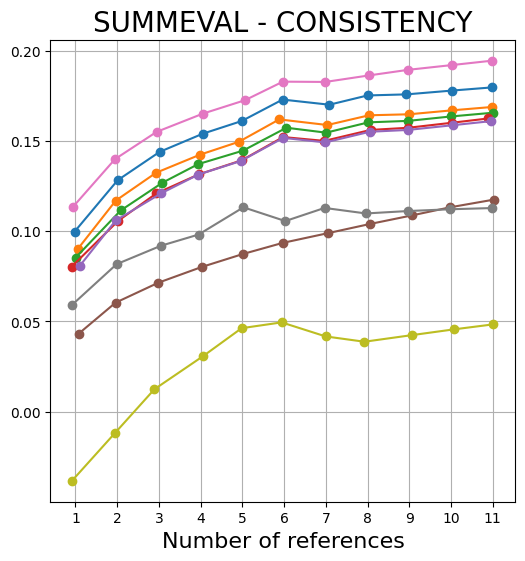}
    \includegraphics[height=4cm]{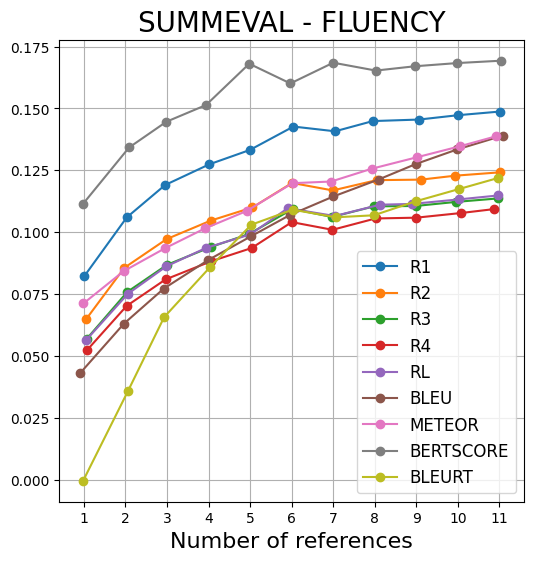}
    \includegraphics[height=4cm]{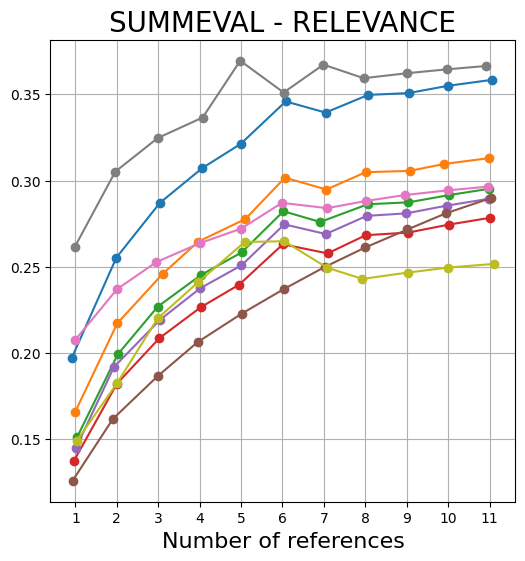}
    \includegraphics[height=4cm]{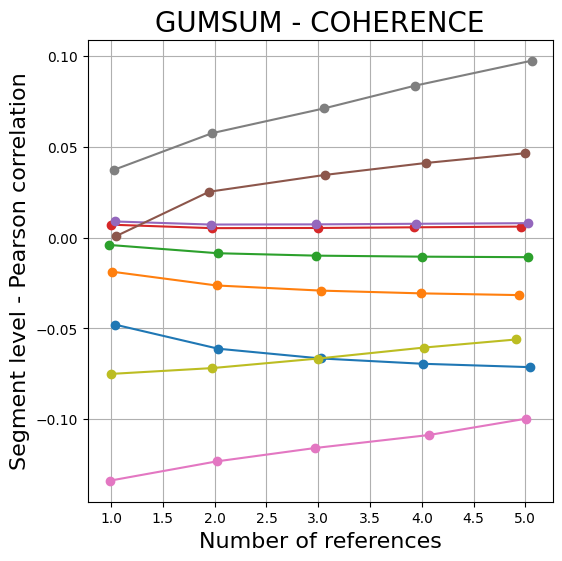}
    \includegraphics[height=4cm]{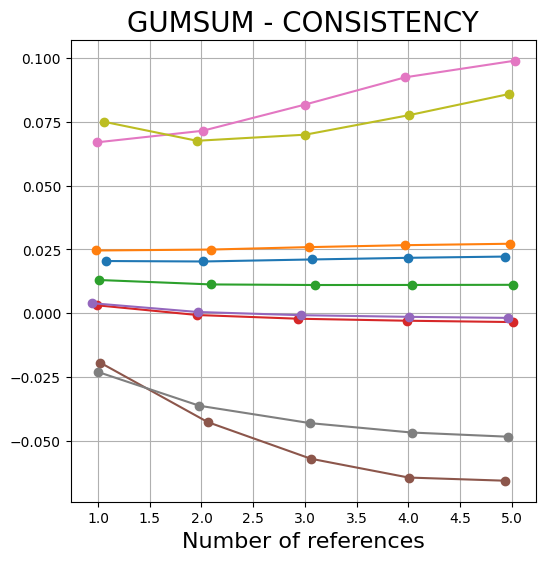}
    \includegraphics[height=4cm]{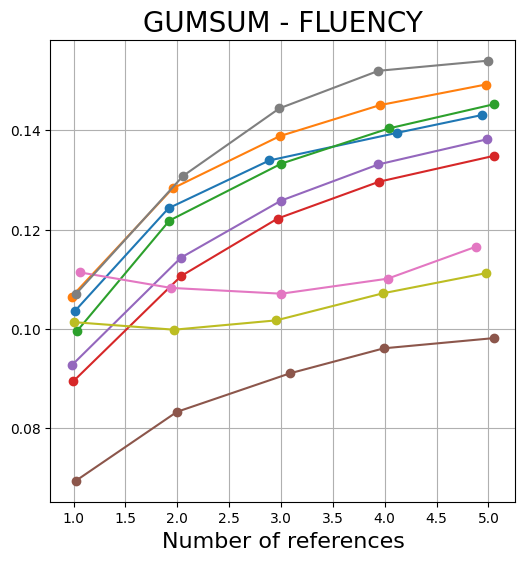}
    \includegraphics[height=4cm]{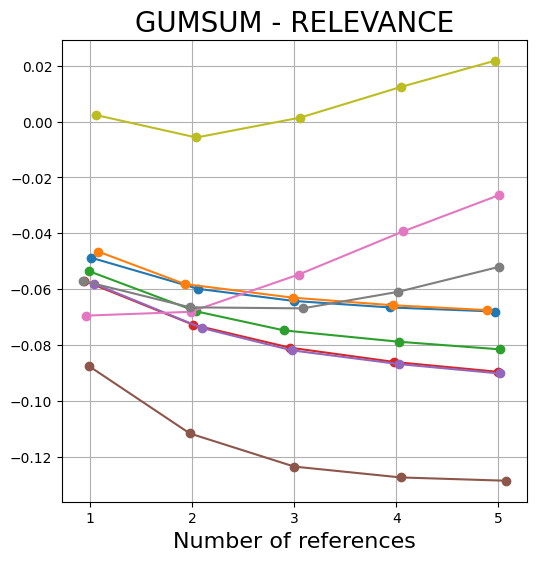}
    \caption{Pearson correlation at the instance level on SummEval (top) and GUMSum (bottom) using ROUGE\textsubscript{mean}.}
    \label{fig:corr_summ_instance_mean}
\end{figure*}

\begin{figure*}[!t]
    \centering    
    \includegraphics[height=4cm]{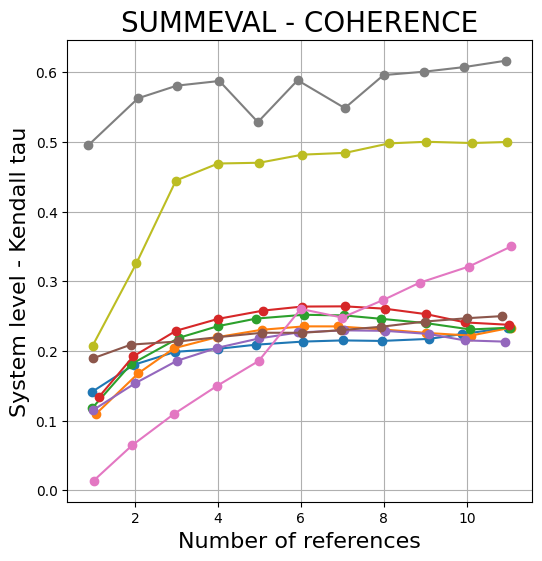}
    \includegraphics[height=4cm]{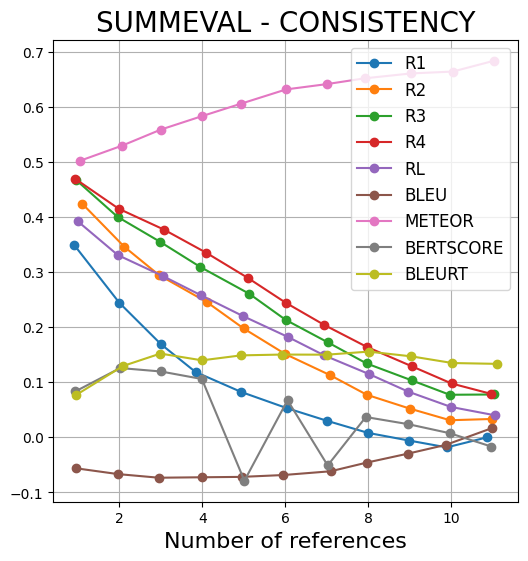}
    \includegraphics[height=4cm]{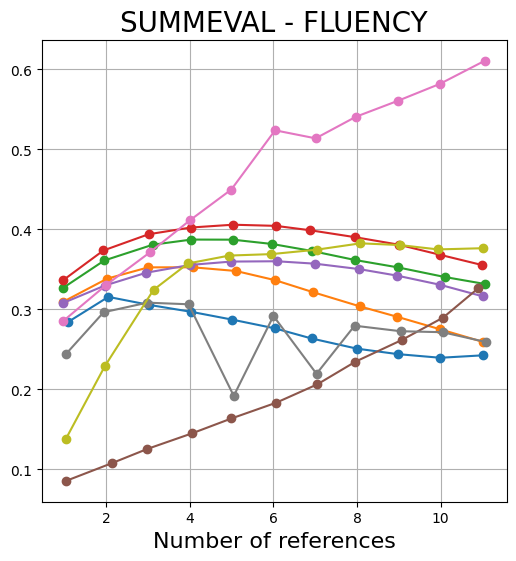}
    \includegraphics[height=4cm]{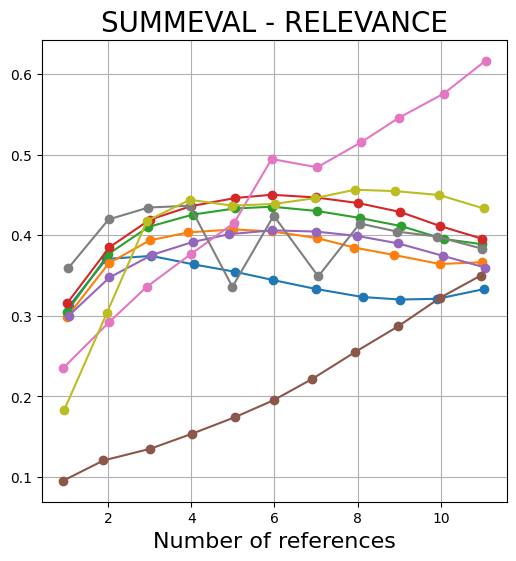}
    \includegraphics[height=4cm]{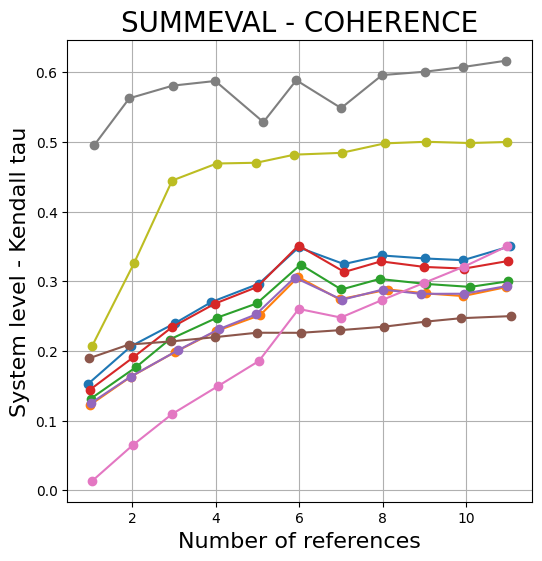}
    \includegraphics[height=4cm]{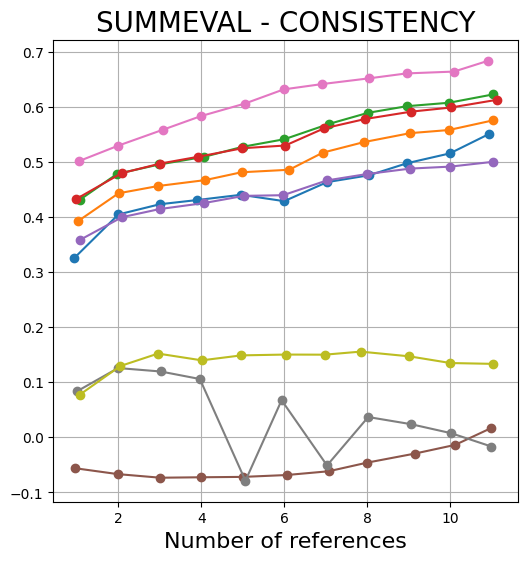}
    \includegraphics[height=4cm]{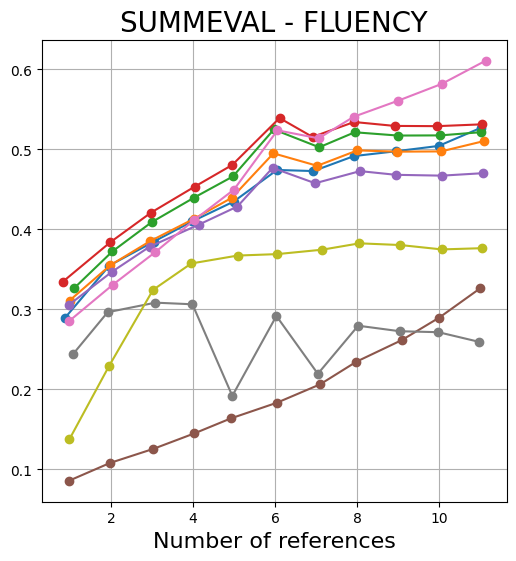}
    \includegraphics[height=4cm]{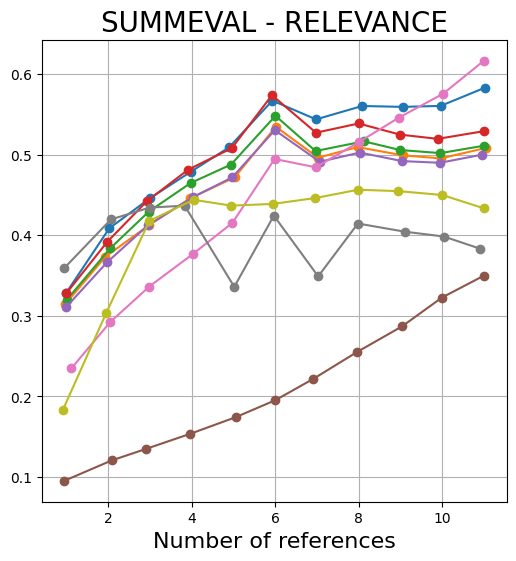}
    \caption{Kendall tau at the system level on SummEval using ROUGE\textsubscript{max} (top) and ROUGE\textsubscript{mean} (bottom).}
    \label{fig:corr_summ_sys}
\end{figure*}

\autoref{fig:corr_summ_instance_mean} shows the instance-level correlation for SummEval (top) and GUMSum (bottom) using ROUGE\textsubscript{avg}.
\autoref{fig:corr_summ_sys} show the system-level correlation on SummEval with ROUGE\textsubscript{max} (top) and ROUGE\textsubscript{avg} (bottom).

\end{document}